\def\year{2020}\relax
\documentclass[letterpaper]{article} % DO NOT CHANGE THIS
\usepackage{aaai20}  % DO NOT CHANGE THIS
\usepackage{times}  % DO NOT CHANGE THIS
\usepackage{helvet} % DO NOT CHANGE THIS
\usepackage{courier}  % DO NOT CHANGE THIS
\usepackage[hyphens]{url}  % DO NOT CHANGE THIS
\usepackage{graphicx} % DO NOT CHANGE THIS
\usepackage{graphicx}  % DO NOT CHANGE THIS
\usepackage{amsmath}
\usepackage{amssymb}

\usepackage{comment}
\usepackage{array}
\usepackage{cellspace}
\usepackage{multirow}
\usepackage{makecell}

\usepackage{booktabs}
\usepackage[table, x11names]{xcolor}
\usepackage{hhline}
\usepackage{placeins}

\newcommand{\etal}{\textit{et al}.}
\newcommand{\cmmnt}[1]{\ignorespaces}
\newcommand*{\Scale}[2][4]{\scalebox{#1}{$#2$}}%

\title{MGD-GAN: Text-to-Pedestrian generation through Multi-Grained Discrimination}
\author{
Shengyu Zhang$^{\dag}$, Donghui Wang$^{\dag}$, Zhou Zhao$^{\dag}$, Siliang Tang$^{\dag}$, Di Xie$^{\star}$, Fei Wu$^{\dag}$\\
$^{\dag}$College of Computer Science and Technology, Zhejiang University, China\\
$^{\star}$Hikvision Research Institute, China\\
\{sy\_zhang, dhwang, zhaozhou, siliang, wufei\}@zju.edu.cn, xiedi@hikvision.com\\
}
\begin{document}

\maketitle

\begin{abstract}
In this paper, we investigate the problem of text-to-pedestrian synthesis, which has many potential applications in art, design, and video surveillance. Existing methods for text-to-bird/flower synthesis are still far from solving this fine-grained image generation problem, due to the complex structure and heterogeneous appearance that the pedestrians naturally take on. To this end, we propose the Multi-Grained Discrimination enhanced Generative Adversarial Network, that capitalizes a human-part-based Discriminator (HPD) and a self-cross-attended (SCA) global Discriminator in order to capture the coherence of the complex body structure. A fined-grained word-level attention mechanism is employed in the HPD module to enforce diversified appearance and vivid details. In addition, two pedestrian generation metrics, named Pose Score and Pose Variance, are devised to evaluate the generation quality and diversity, respectively. We conduct extensive experiments and ablation studies on the caption-annotated pedestrian dataset, CUHK Person Description Dataset. The substantial improvement over the various metrics demonstrates the efficacy of MGD-GAN on the text-to-pedestrian synthesis scenario.
\end{abstract}

\section{Introduction} \label{introduction}

%\paragraph{Introduction + Related Works = 2 pages}
%\paragraph{Methods = 2 pages}
%\paragraph{Experiments = 4pages}

Synthesizing visually authentic images from textual descriptions is a representative topic of creative AI systems. It requires a high-level understanding of natural language descriptions, usually vague and incomplete, and the imaginative ability in order to draw visual scenes. Recently, methods based on deep generative models, e.g., Generative Adversarial Networks \cite{goodfellow2014generative}, have accomplished promising outcomes in this field. With the generation quality increasing, more challenging objectives are expected, i.e., fine-grained image generation\cite{wei2019deep}.

%Automatic synthesis of realistic images within a fine-grained category would be interesting and useful, but current AI systems are still far from this goal.

Fine-grained image generation, as the term implies, focuses on fine-grained categories, such as faces of a certain individual or objects within a subcategory. Existing methods for fined-grained text-to-image generation only focuses on birds and flowers \cite{reed2016generative,Zhang:2017wo,Xu:2018wgb,zhu2019dm,qiao2019mirrorgan}. However, text-to-pedestrian generation is a new research direction and has many potential applications in movie making \cite{ma2017pose}, art creation, and video surveillance — for instance, the appropriate generation of one suspect's portrait according to a witness or victim descriptions. We have carried out an experiment to demonstrate that pedestrian generation can significantly improve the text-to-person search results. Also, text-to-pedestrian generation is a rather difficult problem due to the nature of larger intra-class variances in person generation. The intra-class variances can be summarized as the inter-person variances, such as gender, age, height, and the intra-person variances, such as appearance variances, pose variances.
%We have carried out an experiment to demonstrate that pedestrian generation can significantly improve the text-to-person search results.

\begin{figure}[!t]
\centering
\includegraphics[width=\columnwidth]{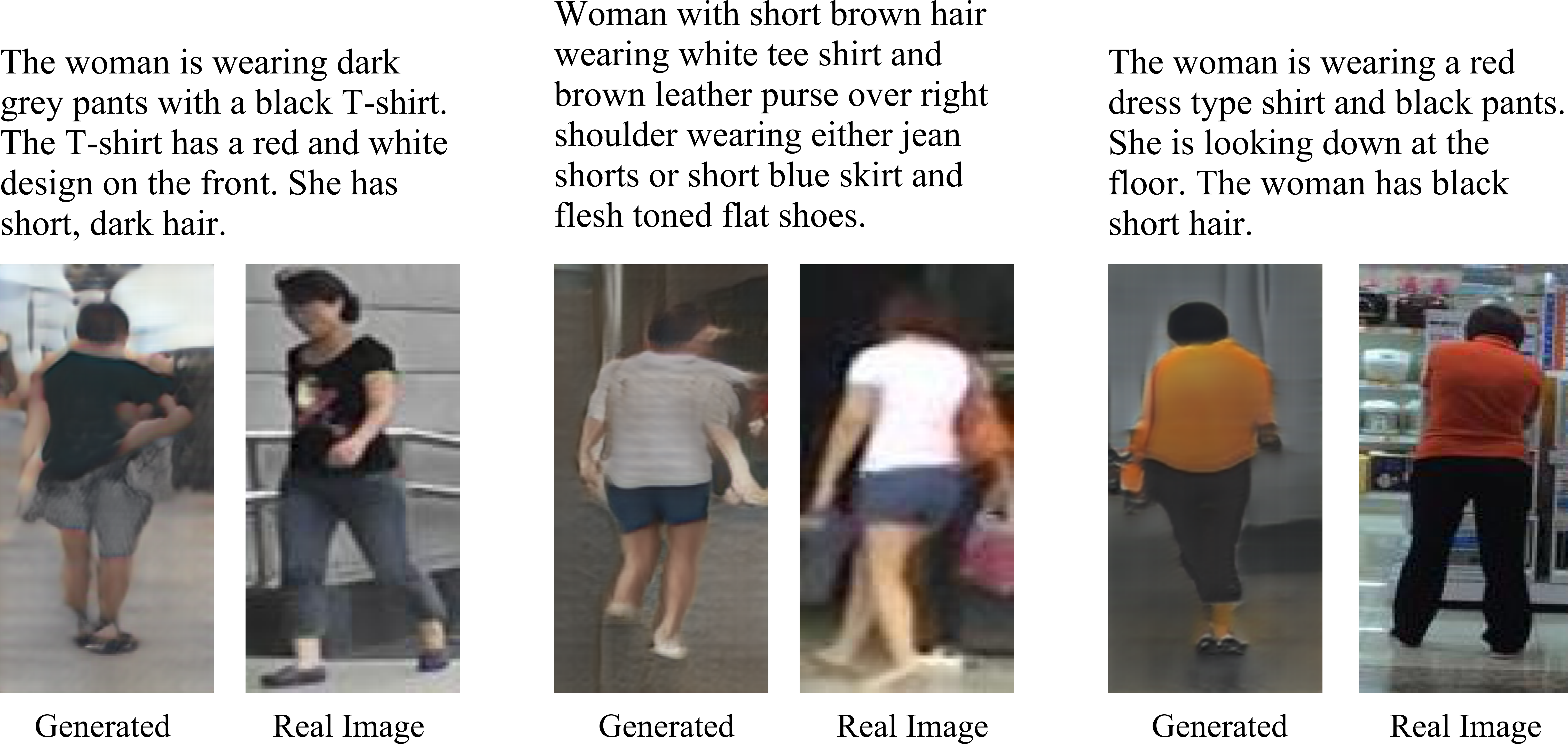}
\caption{
Examples generated by our proposed MGD-GAN.
}
\label{fig:overview}
\end{figure}

%Text-to-pedestrian synthesis techniques have many real-world applications. For instance, in the automated surveillance scenario, when we only have the recorded words from witnesses and want to find potential targets, text-to-pedestrian synthesis can help bridge the gap between the vague natural language descriptions and a reasonable imagination. We have carried out an experiment to demonstrate that pedestrian generation can significantly improve the person search results.

% We have conducted an experiment to reveal that pedestrian generation can improve person search result by a large margin.

%(appearance variances, pose variances, age height etc..)

%To the best of our knowledge, this is first paper that advocates to address the text-to-pedestrian synthesis problem.

Not surprisingly, existing methods that are designed for a general text-to-image synthesis problem are still far from solving the pedestrian generation problem. 
On the one hand, existing methods fail to leverage the task-specific prior knowledge in the pedestrian generation process. To this end, we investigate the person re-id techniques and propose a human-part-based Discriminator, named HPD, which independently penalizes imperfect structure at each human part. 
On the other hand, as stated above, we view text-to-pedestrian generation as a fined-grained generation problem. Although there are works \cite{Xu:2018wgb,zhu2019dm} that exploits a fine-grained word-level attention mechanism in the generation process, the absence of fine-grained discrimination may induce an unbalanced competition between the Generator and Discriminator since they have different levels of abilities. In essence, the Generator focuses on more fine-grained local details while the Discriminator can only capture the global coherence of the image.

To handle the problems above, we enhance the human-part-based Discriminator with a Visual-Semantic Attention module, named VISA. The difference of a regular conditional discriminator and a VISA enhanced discriminator is that rather than signifies the concatenated input of a fake image and a sentence "real" or "fake", whereas the VISA enhanced discriminator scores the importance of each word for each image region and signifies each region "real" or "fake" concerning different words. This setting can help discriminator better penalize the inconsistency between textual descriptions and generated images in a words-regions level, which makes VISA-HPD a local discriminator. As a counterpart of the local one, we construct another self-cross-attended global discriminator functioning at the sentence-image level, penalizing irregular body structure as a whole. %观察整个图片和句子级别.

To the best of our knowledge, our work is the initiative to do text-to-pedestrian synthesis. The most commonly used evaluation metric for the generic image generation, the Inception Score, relies on a classification model. However, there are no specific sub-categories for pedestrian, and thus an Inception Score model trained on a generic image recognition dataset \cite{DBLP:journals/corr/abs-1904-05118} may fail to capture the quality and diversity of pedestrian generation results. To this end, we propose two evaluation metrics, named Pose Score and Pose Variance, grading the generation quality and generation variety simultaneously. Based on a Pose Estimation model pre-trained on a larger pedestrian dataset, the evaluation is easy to compute in practice.% 对于 single-category text-to-image synthesis，用于计算IS的Inception Model通常是在相应数据集的分类任务上训练的，在对于人的质量和多样性评价上有局限性。因此我们提出了 Pose Score 和 Pose Variance 评价指标分别用于行人生成的质量和多样性的评价。

%\subsection{Contribution}

To summarize, this paper makes the following key contributions:
\begin{itemize}
	\item We extend previous fine-grained text-to-image synthesis research by advocating a more challenged pedestrian generation problem.
	\item We proposed a novel MGD-GAN model that exploits a visual-semantic attention enhanced human-part-based Discriminator as well as a self-cross-attended global Discriminator, alleviating the unbalance competition bottleneck between Discriminator and Generator.
	\item We present two novel pedestrian generation metrics, named pose score and pose variance, as a non-trivial complement to existing ones.
	\item The proposed MGD-GAN model achieves the best results on the challenging CUHK-PEDES dataset. The extensive ablation studies verify the effectiveness of each component within the MGD-GAN model in different aspects, i.e., generation quality, generation variety, and the consistency with the textual description.
\end{itemize}

\section{Related Works}

\subsection{Person image generation}
%
%Previous person image generation methods mainly focus on the pose-transfer task, which aims to generate person images with the expected poses. For instance, the $PG^2$ model \cite{ma2017pose} transfers the condition image, and the target pose into an initial but coarse image and refines that image in an adversarial way. Grigorev \etal \cite{grigorev2019coordinate} estimates of the complete body surface texture and establish the correspondence between the source and the target view even when the pose change is drastic. Recently, Zhou \etal \cite{DBLP:journals/corr/abs-1904-05118} present a two-stage model to manipulate the visual appearance (pose and attribute) of a person image according to natural language descriptions. Despite this progress, our work synthesizes a person image based solely on the textual descriptions. This is a more challenging task and has real-world applications, as stated in the \textit{Introduction} section.
%
Previous person image generation methods mainly focus on the pose-transfer task, which aims to generate person images with the expected poses \cite{ma2017pose,ma2018disentangled,siarohin2018deformable,grigorev2019coordinate,zhu2019progressive}. Recently, Zhou \etal \cite{DBLP:journals/corr/abs-1904-05118} presents a two-stage model to manipulate the visual appearance (pose and attribute) of a person image according to natural language descriptions. Despite this progress, our work synthesizes a person image based solely on the textual descriptions, which is a more challenging task and has different applications.

%- PG2 Liqian Ma, Jia Xu, Qianru Sun, Bernt Schiele, Tinne Tuytelaars,
%and Luc Van Gool. Pose guided person image generation.
%In NIPS, 2017.
%- Balakrishnan et al. [1] present a modular generative
%network which separates a scene into various layers
%and moves the body parts to the desired pose. Guha Balakrishnan, Amy Zhao, Adrian V Dalca, Fredo Durand, and John Guttag. Synthesizing images of humans in unseen poses. In CVPR, 2018. 1, 2
%- Esser et al. [5] present a conditional U-Net shapeguided
%image generator based on VAE for person image
%generation and transfer. It is desirable to edit and manipulate
%person images according to natural language descriptions. Patrick Esser, Ekaterina Sutter, and Bjorn Ommer. A variational
%u-net for conditional appearance and shape generation.
%In CVPR, 2018. 1, 2
%- Text guided Image Synthesis

\subsection{Text-to-image generation}

The challenging and open-ended nature of text-to-image generation lends itself to a variety of diverse models. There are mainly two lines of works. The first is generating complex scenes with multiple objects \cite{hong2018inferring,li2019object,hinz2019generating}. The other is the generation on fine-grained categories \cite{wei2019deep}. In this line of works, GAN-INT-CLS \cite{reed2016generative} firstly learns a joint embedding of text and images to capture the fine-grained relationship between them. After that, a conditional GAN is introduced to learn the mapping from text embeddings to images. Zhang \etal \cite{Zhang:2017wo} stacks multiple generators to remedy the fine-grained details from previous generators. HD-GAN \cite{zhang2018photographic} uses hierarchically-nested discriminators to generate photographic images. More recently, the AttnGAN \cite{Xu:2018wgb} can synthesize fine-grained details of subtle regions by performing an attention equipped generative network. The main difference between our method and previous arts is two-fold. First, we advocate solving a more challenging text-to-pedestrian generation problem besides the existing birds\//flowers generation. Second, we present the global-local attentional discriminators to ease the unbalanced adversarial training problem within previous works.

\begin{figure*}[t] \begin{center}
    % Figure 2 (Figure 5)
    % NOTE: "samples-box-mask-image.pdf" are automatically generated from a script:
    \includegraphics[width=0.9\textwidth]{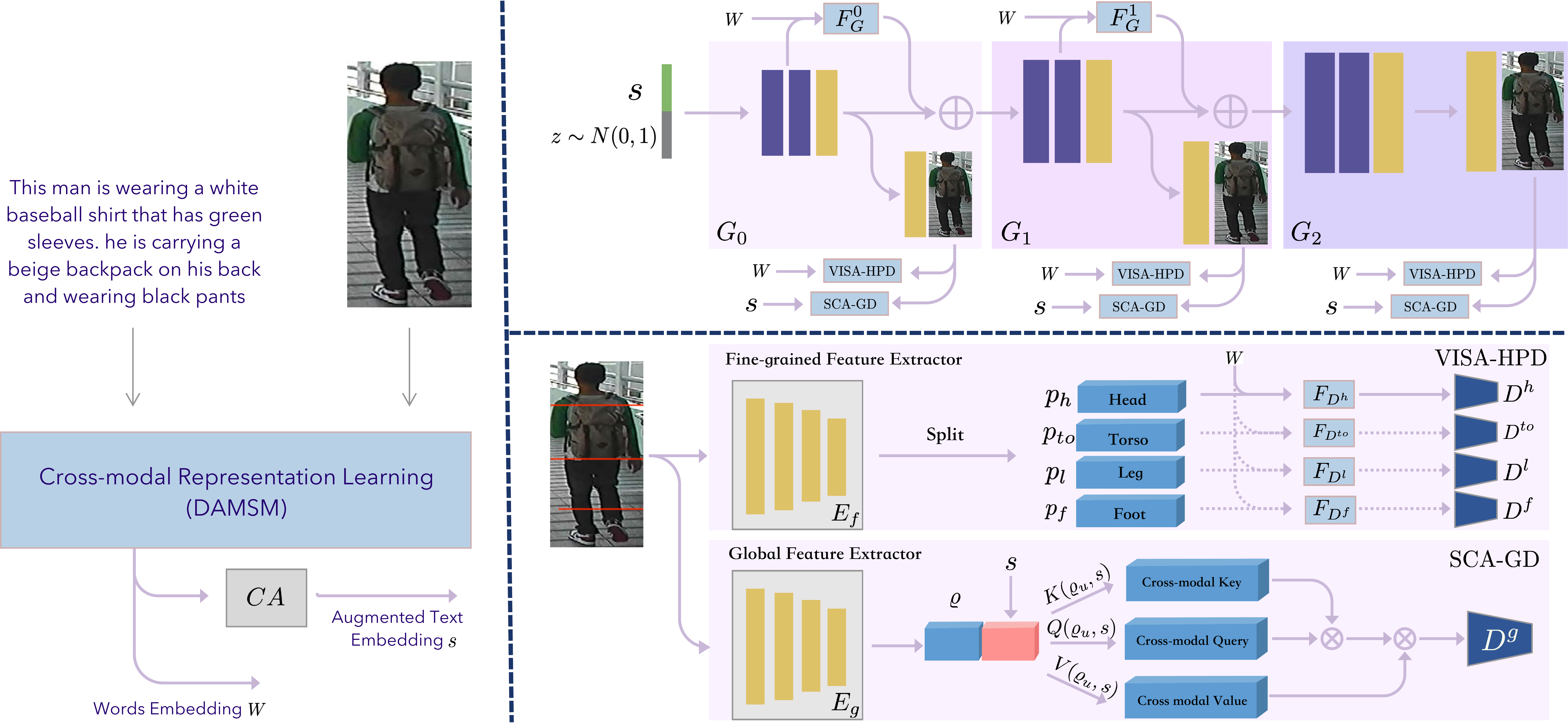}
    \caption{
    The overall scheme of our proposed MGD-GAN model. The initial text embedding $s$ and words embeddings $W$ are obtained by a pre-trained cross-modal representation learning model named DAMSM. At each training stage $i$, we build a human-part-based discriminator VISA-HPD functioning at the words-regions level and a global discriminator SCA-GD at the sentence-image level, as well as a fine-grained generator.
}
\label{fig:structure}
\end{center} \end{figure*}
%----------------------------figure end-------------------------

%CVAE-GAN \cite{bao2017cvae} was an initiative to tackle this problem, which combines a variational auto-encoder with the GANs.

%Concerning the unconditional image generation, the BigGANs \cite{brock2018large} firstly train Generative Adversarial Networks at the largest scale yet attempted, and study the instabilities specific to such scale. More recently, based on BigGANs, Donahue, et al. \cite{donahue2019large} present the BigBiGAN model by adding an encoder and modifying the discriminator, which achieves the state of the art in unsupervised representation learning as well as in unconditional image generation.

%In terms of conditional image generation, TAGAN \cite{nam2018text} generates semantically manipulated images while preserving text-irrelevant contents by adding a text-adaptive discriminator. To perform image-to-image translations for multiple domains using only a single model, Choi, et. al. \cite{choi2018stargan} 

%- unconditional image generation (bigGAN, SAGAN)
%- Image Editing (Dong, TAGAN)
%- Style Transfer (CycleGAN, )
%- Image Inpainitng (Pluristic Image Inpainting)

%- Text-to-Image synthesis (Reed, StackGAN, AttnGAN, vmCAN, ObjGAN)

% 最后重点提一下和 text guided image synthesis 的差别

\section{Multi-Grained Discrimination enhanced Generative Adversarial Network}

In this section, we will elaborate on each building block comprising the MGD-GAN model. As shown in Figure \ref{fig:structure}, MGD-GAN mainly embodies three modules: the multi-stage attentional generator, the visual-semantic attention enhanced human-part-based discriminator, and the self-cross-attended global discriminator. %Details of these modules will be clarified in the next subsections.

Technically, given the sentence feature $s$ and words feature $W$, we want to synthesize the desired human picture $I$. The sentence feature $s \in \mathbb{R}^{N_s} $ and the words feature $W =\{w_1,w_2,...,w_T\},w_t\in \mathbb{R}^{N_w}$ are pre-trained using a cross-modal representation learning network named DAMSM \cite{Xu:2018wgb}. $T$ is the specified max sequence length of sentences, $N_s$ is the dimension of sentence features and $N_w$ is the dimension of words features.

% However, low-resolution pictures generally lack sharp image components and could involve shape distortions. Some semantic information within the input text $s$ could also be overlooked.

\subsection{Multi-stage generation strategy}
We follow a standard multi-stage baseline \cite{zhang2017stackgan++,zhu2019dm} to develop our generator. Rather than creating a high-resolution picture directly from the textual description $s$, we simplify the generation process by first creating a low-resolution picture $I_0$, which relies on only rough shapes and colors. To remedy fine-grained errors, we incorporate a refinement generator $G_{i+1}$ which is conditioned on the low-resolution outcomes $I_{i}$ and the words features $W$. The first generator $G_0$ and multiple refinement generators $\{ G_1, ..., G_{m-1} \}$ can be stacked sequentially and form the multi-stage generation process.

In order to enforce the model to produce fine-grained details within the words feature \cite{Xu:2018wgb} and carry out compositional modeling \cite{cheng2018sequential} , we construct several attention modules $F^{i}_{G}$ for adjacent generators $ (G_{i},G_{i+1}) $. The structure of $F^{i}_{G}$ is the same as VISA.
%To carry out compositional modeling, the design is forced to produce areas and related characteristics in accordance with the text description

%\subsection{Human-part-based Discriminator}

\subsection{VISA-HPD: Visual-Semantic Attention enhanced Human-Part-based Discriminator} \label{VISA-HPD}

To leverage some domain knowledge for the fine-grained text-to-pedestrian problem, we investigate some popular methods used in person re-identification research field and propose a human-part-based approach. The core idea in the behind is simple but effective. In order to tackle the complex structure of the human body, we split the encoded feature $\rho$ from the generated image $I$ vertically and equally into several parts, i.e., the head part $p_h$, the torso part $p_{to}$, the leg part $p_l$ and the foot part $p_f$. The correctness, the coherence, and the faithfulness to the corresponding words and phrases of each human part are individually scored by a related discriminator, i.e., $ D^{ \kappa }, \kappa \in \{ h,to,l,f \} $. We note that human part discriminators share a common fine-grained feature encoder $E_f$ but do not share parameters.

Previous GAN-based models for fine-grained text-to-image generation typically equip the Generator with attention mechanisms. In order to grade the relevance of regions and words, we develop a Visual-Semantic Attention, named VISA, for each human part discriminator. The VISA module $F_{D^ \kappa }$ takes the words features $W \in \mathbb{R}^{N_w \times T }$ and the human part regions features $p_{\kappa} \in \mathbb{R}^{N_r \times N/4}$ as input. $N$ is the number of regions in $\rho$ and $N_r$ is the dimension of regions.

% in the intermediate feature map and

% However, there is a dearth of fine-grained discrimination in the zero-sum game during training. To this end, we develop a Visual-Semantic Attention for each human part discriminator. In order to capture fine-grained details in the words-regions level, our VISA module $F_D$ takes the words features $W \in \mathbb{R}^{N_w \times T }$ and the encoded regions features $h = E_{D}(I) \in \mathbb{R}^{N_r \times N}$ as input. $N$ is the number of regions in the output feature map and $N_r$ is the dimension of regions.

For each image region feature $\rho_u$, we compute the attention weights with words features $W$ using a Fully-Connected layer $f$ and a softmax layer. We obtain the final words-attended region feature as the addition of original region feature and the weighted sum of words features.

\begin{align}
	\alpha_{u,t} &= \frac{\exp(f(\rho_u, w_t))}{\sum_{t}\exp(f(\rho_u, w_t))} \\
%	\alpha_{ut} &= softmax(f(\rho_u, w_t)) \\
	\rho_u' &= \sum_t{\alpha_{u,t} w_t} \\
	r_u &= \rho_u + \rho_u'
\end{align}

%The VISA formulation is similar to that of AttnGAN \cite{Xu:2018wgb}. However, we are driven by different motivations and there are several crucial differences. First, we use additive attention instead of dot-product attention used by AttnGAN. Second, we use the addition of attended words feature and image feature for efficient inference. Third, we use VISA to obtain fine-grained discrimination and consequently achieve a balanced and robust adversarial training.

Hence, the discrimination process for generated image $I$ can be expressed as:

\begin{align}
	& \rho = E_{f}(I) \\
	& p_h, p_{to}, p_l, p_f = Split(\rho) \\
	& y_{\kappa} = D^{\kappa}(F_{D^ \kappa }(W, p_{\kappa}))
\end{align}

where $ y_{\kappa} \in (0,1)$ is the output of each human part discriminator, which grades the generation quality and faithfulness to attended words of each human part.

\subsection{Self-cross-attended Global Discriminator}

As a non-trivial counterpart of the local discriminator stated above, a self-cross-attended global discriminator is proposed to capture the coherence of the body structure as a whole. Extending beyond the Self-Attention GAN \cite{zhang2018self}, we not only build a self-attention map of spatial context but also harness cross-modal context at the sentence-image level. Specifically, we build the self-cross-attention scores of cross-modal context by:

\begin{align}
	\beta_{v, u} = \frac{\exp(c_{uv})}{\sum^N_{u=1}{\exp(c_{uv})} },\text{where }  c_{uv} = K(\varrho_u, s)^TQ(\varrho_v, s)
\end{align}

where $\varrho_u, \varrho_v \in \mathbb{R}^{N_r} $ denote region features within $ \varrho = E_g(I) $ extracted by a global feature extractor $E_g$. $K$ and $Q$ are joint image-sentence feature spaces, formulated as.

\begin{align}
	K(\varrho_u, s) = W_k([\varrho_u, s])
\end{align}

 Therefore, $\beta_{v, u}$ indicates to what magnitude the $u^{th}$ image region is attended to synthesize the $v^{th}$ image region concerning the semantic context at the sentence level. The final self-cross-attention map $o = (o_1, o_2, ..., o_v, ..., o_N) \in \mathbb{R}^{ N_r \times N } $ can be obtained by:
 
\begin{align}
	o_v = W_z(\sum_{u=1}^N \beta_{v,u} V(\varrho_u, s) )
\end{align}

where $V$ is another 1x1 convolution besides $K$ and $Q$.

\subsection{Objective Functions}

The objective function of our MGD-GAN is composed of three parts, i.e., the adversarial loss, the conditioning augmentation loss, and the DAMSM loss.

\paragraph{Adversarial Loss} A common practice is to employ two adversarial losses: the unconditional adversarial loss and the conditional visual-semantic adversarial loss. We further expand the conditional visual-semantic adversarial loss hierarchically, i.e., a fine-grained words-regions adversarial loss and a global sentence-image adversarial loss.

The generator $G$ and discriminator $D$ are alternatively trained at each learning stage of MGD-GAN. In particular, the adversarial loss of the generator $G_i$ at the $i^{th}$ stage can be defined as:

\begin{equation}
	\begin{aligned}
	\mathcal{L}_{G_i}^{adv} &= -1/3 [ \underbrace{ \mathbb{E}_{I_g \sim p_{G_i}} \log{ D_i^g(I_g) } }_{ \text{unconditional loss} } + \underbrace{ \mathbb{E}_{I_g \sim p_{G_i}} \log{ D_i^g(I_g, s) } }_{ \text{ global conditional loss } } \\ &+ \underbrace{ 1/4 \sum_{ \kappa }{ \mathbb{E}_{I_g \sim p_{G_i}} \log{ D_i^{ \kappa }(I_g, W) }  } }_{ \text{ local conditional loss } } ]
	\end{aligned}
\end{equation}

where the $D_i^g$ stands for the self-cross-attended global discriminator. The unconditional loss is intended to distinguish the generated image from the real image. The global conditional loss and local conditional loss are designed to determine whether the generated image is faithful to the input description at the sentence-image level and the words-regions level, respectively.

We define the loss function of two discriminators as:

\begin{equation}
	\begin{aligned}
	\Scale[0.95]{\mathcal{L}_{D_i^g}} &= \Scale[0.85]{-\frac{1}{2} [ \mathbb{E}_{ I_{d} \sim p_{data} } \log{ D_i^g(I_{d}) } + \mathbb{E}_{ I_{g} \sim p_{G_i} } \log{ (1 - D_i^g(I_{g})) }} \\
	&\Scale[0.95]{+ \mathbb{E}_{ I_{d} \sim p_{data} } \log{ D_i^g(I_{d}, s) } + \mathbb{E}_{ I_{g} \sim p_{G_i} } \log{ (1 - D_i^g(I_{g}, s)) }]}
	\end{aligned}
\end{equation}

\begin{equation}
	\begin{aligned}
	\Scale[0.9]{\mathcal{L}_{D_i^{ \kappa }}} &= 
	\Scale[0.85]{- \mathbb{E}_{ I_d \sim p_{data} } \log{ D_i^{\kappa}(I_d, W) } - \mathbb{E}_{ I_g \sim p_{G_i} } \log{ (1 - D_i^{\kappa}(I_g, W)) }} 
	\end{aligned}
\end{equation}

It is worth noting that the loss function of the human-part discriminator $\mathcal{L}_{D_i^{\kappa}}$ does not contain an unconditional adversarial loss. We assume that a typical conditional adversarial loss does measure the visual quality of images as well as the visual-semantic correspondence and that the body structure is better captured as a whole. %our global discriminator has already captured the coherence of the body structure as a whole.

\paragraph{Conditioning Augmentation Loss} To mitigate the discontinuity problem caused by limited training data, we corporate a regularization term, named Conditioning Augmentation \cite{Zhang:2017wo}, formulated as follows:

\begin{equation}
	\begin{aligned}
		\mathcal{L}_{cond} = D_{KL}( \mathcal{N} ( \mu(s), \Sigma(s)  ) || \mathcal{N}(0, \mathrm{I}) )
	\end{aligned}
\end{equation}

where $D_{KL}$ represents the Kullback-Leibler divergence. The estimation of mean $ \mu(s) $ and diagonal covariance $ \Sigma(s) $ are modeled by fully-connected layers.

%Mathematically, the regularization term is the Kullback-Leibler divergence between a standard Gaussian distribution and an estimated Gaussian distribution of the training sentences. The estimation of mean $ \mu(s) $ and diagonal covariance $ \Sigma(s) $ are modeled by fully-connected layers.

\paragraph{DAMSM Loss} As a common practice \cite{Xu:2018wgb,cheng2018sequential,zhu2019dm} ,  we employ a cross-modal representation learning based module named DAMSM. This module provides the initial sentence and words embeddings, as well as a matching loss $\mathcal{L}_{DAMSM}$ of the generated image and conditioned text. %DAMSM is designed to warm-start the generation training and enforce the generation correspondence alongside the conditional adversarial losses.

Therefore, the final loss function of the generator can be written as:

\begin{equation}
	\begin{aligned}
		\mathcal{L}_{G_i} = \sum_i{ \mathcal{L}_{G_i}^{adv} + \lambda_1\mathcal{L}_{cond} + \lambda_2\mathcal{L}_{DAMSM} }
	\end{aligned}
\end{equation}

%--------------------------------table---------------------
\begin{table*}[t]
%\begin{strip}
\centering
\setlength\doublerulesep{0.5pt}
%\footnotesize  
\small
\begin{tabular}{l|ccccccc}
%\hline
%\toprule
& \multicolumn{7}{c}{CAPTION GENERATION}  \\
%\cline{1-9}
% <<<
\multicolumn{1}{c|}{Method}                 & BLEU-1         & BLEU-2         & BLEU-3         & BLEU-4         & METEOR   &ROUGE\_L          & CIDEr          \\
%\midrule
\hline \hline
Reed \etal \cite{reed2016generative}            & 0.496          & 0.321          & 0.218          & 0.151          & 0.211          & 0.454           & 0.949          \\
StackGAN   \cite{Zhang:2017wo}  & 0.512   & 0.337   & 0.233   & 0.163   & 0.219   & 0.468   & 1.125 \\
AttnGAN   \cite{Xu:2018wgb}     &   0.561   & 0.396   & 0.293   & 0.222   & 0.253   & 0.525   & 1.519
        \\
\hline \hline
\iffalse\rowcolor{SeaGreen1!20!}\fi Ours   & \textbf{0.565} & \textbf{0.401} & \textbf{0.299} & \textbf{0.229} & \textbf{0.257} & \textbf{0.530}  & \textbf{1.553} \\
%\midrule
%\hline 
%Baseline  & 0.556          & 0.353          & 0.219          & 0.139          & 0.162          & 0.400           & 11.94 $\pm$ 0.09          \\ 
%Baseline2   & 0.573          & 0.373          & 0.239          & 0.156          & 0.169          & 0.440           & 12.40 $\pm$ 0.08          \\
\hline
Real images (upper bound)   & 0.599    & 0.443    & 0.340    & 0.268    & 0.282    & 0.565     & 1.855
           \\
\hline
% >>>
\end{tabular}
%\medskip
\caption{
    Quantitative evaluation results of caption generation on different natural language generation metrics.
%    The last row presents the caption generation performance 
%    on real images, which corresponds to upper-bound of caption generation metric.
    The last row presents the performance of real images, which can be viewed as the upper-bound of the evaluation metrics.
    %
    %Higher is better in all columns. %\wook{Reduce words bit}
    %
}
\label{tab:eval_caption}
\end{table*}

\section{Pose Score \& Pose Variance}
In current literature, the numerical assessment approaches for GANs are not explicitly intended for pedestrian generation, such as Inception Score.

Under the observation that pedestrian has no salient sub-categories like birds and flowers, the classification model for Inception Score trained on a generic image recognition dataset may fail to evaluate the generation quality and diversity competently. To this end, we propose two pedestrian-specific generation metrics named Pose Score and Pose Variance as a non-trivial complement to the Inception Score.

Specifically, we first pre-train a pose estimation model \cite{cao2017realtime} on a larger COCO 2016 keypoints challenge dataset \cite{lin2014microsoft}. For each generated pedestrian $I_{ \epsilon }, \epsilon \in \{ 1, ..., \Xi \} $ , we detect the 2D positions of body parts, represented as $ B_{ \epsilon } = \{ b_{\epsilon,1}, b_{\epsilon,2}, ..., b_{ \epsilon,\tau }, ..., b_{\epsilon,J} \} $ where $b_{\epsilon,\tau} \in \mathbb{R}^2$, one per part. Some parts may not be detected because the part is occluded or blurred. Under the assumption that there is no occlusion in the CUHK-PEDES dataset, we use the average ratio of the number of detected parts and the upper-bound, i.e., $ PS = \frac{1}{\Xi} \sum_{ \epsilon } \frac{\#B_{ \epsilon }}{18} $, as the Pose Score.

%Some parts may not be detected, either due to the occlusion of the part, or to the part being blurred. Under the assumption that there is no occlusion of any body part for each pedestrian in the CUHK-PEDES dataset, we use the ratio of the number of detected parts and the number of ground-truth body parts, i.e., $ PS = \frac{\#B}{J} $, as the Pose Score. 

%---------------------------------figure-------------------------

%\begin{figure*}[t] \begin{center}
%    % Figure 2 (Figure 5)
%    % NOTE: "samples-box-mask-image.pdf" are automatically generated from a script:
%    \includegraphics[width=\textwidth]{./figures/colorChange.pdf}
%    
%    \caption{
%    Qualitative results of text2pedestrian generation by our model. We additionally show that our model can synthesize visually authentic images with diversified appearances and human poses by deliberately changing the attribute.
%}
%\label{fig:colorChange}
%\end{center} \end{figure*}

As for Pose Variance, we individually compute the variance of each body part for all generated pedestrians and average them. The intuition is that the larger the variance of human poses is, the more diversified pedestrians the model can generate. Mathematically, the proposed Pose Variance can be computed as:
%Mathematically, let $\phi_{\tau,\iota }$ represents the set of the normalized $\iota^{th}$ axes of the $\tau^{th}$ body part for all generated pedestrians. 
%\phi_{\tau,\iota}

\begin{equation}
	\begin{aligned}
		PV = \exp( \frac{1}{J*2} \sum_{\tau=1}^{J}{\sum_{\iota=1}^{2}{Var( \{ \frac{ b_{\epsilon,\tau,\iota } }{ b_{max} } \}_{ \epsilon=1 }^{ \Xi } )}} )
	\end{aligned}
\end{equation}

where $b_{max}$ is a normalization factor and is set to 256, which is the width or height of generated image $I$.

%We add an exponential operation in order to enlarge the value without affecting the magnitude.

%\subsection{Pose Score \& Pose Variance}

\section{Experiments}

% 最好写点东西

%--------------------------------figure-------------------------

\begin{figure*}[t] \begin{center}
    % Figure 2 (Figure 5)
    % NOTE: "samples-box-mask-image.pdf" are automatically generated from a script:
    \includegraphics[width=\textwidth]{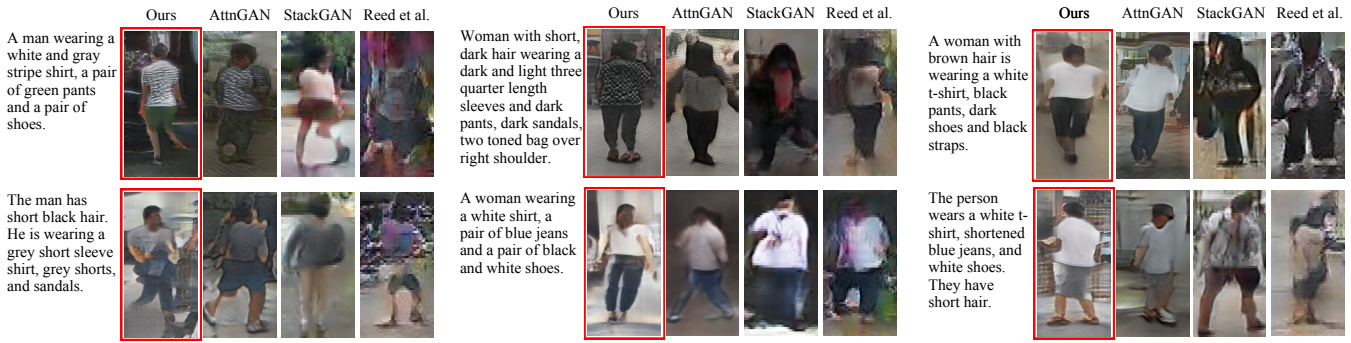}
    \caption{
    Qualitative comparison of four methods. Generated images are conditioned on text descriptions from CUHK-PEDES test set. The image generated by our method is listed in the first column for every example and is boxed off in red. Despite some failure of small objects like straps, our method outperforms other methods qualitatively in most cases.
	}
\label{fig:compare}
\end{center} \end{figure*}

%-----------------------------figure end-----------------------

\subsection{Experiment setup}

\paragraph{Dataset} As far as we know, the CUHK-PEDES dataset \cite{DBLP:conf/cvpr/LiXLZYW17} is the only caption-annotated pedestrian dataset. It contains 40,206 images over 13,003 persons. Images are collected from five existing person re-identification datasets, CUHK03 \cite{li2014deepreid} , Market-1501 \cite{zheng2015person} , SSM \cite{xiao2016end} , VIPER \cite{gray2007evaluating} , and CUHK01 \cite{li2012human} while each image is annotated with 2 text descriptions by crowd-sourcing workers. Sentences incorporate rich details about person appearances, actions, poses. The nature of abundant vocabulary and little repetitive information makes it a better dataset for the fine-grained text-to-image generation. We follow the original dataset split, which includes 34,054 images for training, 3,078 for validation, and 3,074 for testing.    %Table \ref{table:stat} lists the statistics of the dataset. %It is worth noting that multiple person images may belong to the same person, and there is no person-id overlap in our splits.

\paragraph{Evaluation metric}

%\begin{table}[!t]
%\centering
%\setlength{\tabcolsep}{6.5pt}
%\setlength\doublerulesep{0.5pt}
%\begin{tabular}{l|ccc}
%%\toprule
%%Dataset & Metric & AttnGAN & DM-GAN \\ \midrule
%%CUB & FID$\downarrow$ & 23.98 & \textbf{16.09} \\
%% & R-precision$\uparrow$ & 67.82$\pm$4.43 & \textbf{72.31$\pm$0.91}\\ \midrule
%%COCO & FID$\downarrow$ & 35.49 & \textbf{32.64} \\
%% & R-precision$\uparrow$ & 85.47$\pm$3.69 & \textbf{88.56$\pm$0.28}\\
%
%\multirow{2}{*}{Dataset} &   \multicolumn{3}{c}{CUHK-PEDES}        \\ 
%  & train & val & test \\
%\hline  \hline
%
%\#samples  &  34,054   & 3,078 & 3,074 \\
%\hline
%\#person  &  11,003   & 1000 & 1000 \\
%
%\hline
%\end{tabular}
%\caption{
%Statistics of the CUHK-PEDES dataset.
%}
%\label{table:stat}
%\end{table}

%\begin{figure}[!t]
%\centering
%\includegraphics[width=\columnwidth]{./figures/colorChange.pdf}
%%
%\caption{
%%
%Examples generated by our proposed MGD-GAN. Generated images have different poses and details with ground-truth images while preserving most of the fine-grained attributes in text input.
%%
%%
%%
%}
%\label{fig:overview}
%\end{figure}

%IS
%Pose Score \& Pose Variance
%R-precision \cite{satta2013appearance}
%Caption Generation

We choose to evaluate the generation result using various metrics, i.e., Inception Score, our newly proposed Pose Score \& Variance, and Caption Generation.

\textit{Inception Score -- } Following standard practice, we use a pre-trained Inception v3 network to compute the Inception Score \cite{salimans2016improved}. Inception Score measures the visual quality and the generation diversity and has been known to be correlated with human perceptions.

%, which is the KL-divergence between the conditional class distribution and the marginal class distribution

\textit{Pose Score \& Pose Variance -- } Our newly proposed Pose Score/Variance metrics, which separately measure the visual quality of different body parts and generation diversity concerning human poses.

\textit{Caption Generation -- } In order to reflect whether the generated image is well-conditioned on input text, we adopt a caption generation based approach \cite{hong2018inferring}. The intuition behind this approach is that if the generated image is faithful to input text, a well-trained caption model on the same dataset can reconstruct the input text accordingly. We use this caption architecture \cite{xu2015show} and this implementation \footnote{https://github.com/sgrvinod/a-PyTorch-Tutorial-to-Image-Captioning} trained on the CUHK-PEDES dataset. We measure the similarity of generated captions and input text using four standard Natural Language Generation metrics, BLEU \cite{papineni2002bleu}, METEOR \cite{banerjee2005meteor} , ROUGE\_L \cite{lin2004rouge} and CIDEr \cite{vedantam2015cider}.

\paragraph{Implementation Details}
% 参考 DM-GAN，Mirror-GAN
As stated in the previous section, we obtain the sentence embeddings and words embeddings from DAMSM and fix them during training. We have $m=3$ training stages which generate images of resolutions $64 \times 64 $, $128 \times 128 $ and $256 \times 256 $ individually. We set $N_w = 256$ and $ N_r = 512 $ as the dimension of words features and intermediate region features. The hyperparameter of generator loss is set as $ \lambda_1 = 1 $ and $ \lambda_2 = 5 $. The dimension of augmented sentence embedding $N_s$ is set to 100.

%--------------------------------table-------------------------

\begin{table}[!t]
\centering
\setlength{\tabcolsep}{9.5pt}
\setlength\doublerulesep{0.5pt}
\begin{tabular}{l|cccc}

%\cline{1-9}
% <<<

\multicolumn{1}{c|}{Models} &    IS      &   PS    &    PV     \\
\hline \hline

Reed \etal            & 4.32 $\pm$ 0.194          &    0.275        &    1.935                 \\
\hline
StackGAN  & 4.79 $\pm$   0.184   & 0.362    &   1.945  \\
\hline
AttnGAN     &  5.07 $\pm$ 0.396   & 0.465    & 2.039  
        \\
\hline
MGD-GAN & \textbf{5.74 $\pm$ 0.526}    & \textbf{0.489}    & \textbf{2.053}
\\
\hline

\end{tabular}
\caption{
Quantitative evaluation results on generation visual quality. IS, PS and PV stands for the Inception Score and the proposed Pose Score/Variance metrics. The PS and PV of real images are 0.774 and 2.388.
}
\label{table:eval_quality}
\end{table}

%---------------------------table end-------------------------

\subsection{Quantitative evaluation}

As far as we know, we are the initiative to do text-to-pedestrian generation. We compare our method with several works for text-to-bird/flower generation.\cite{reed2016generative,zhang2017stackgan++,Xu:2018wgb}.

%\begin{table}[!t]\setlength{\tabcolsep}{6.5pt}
%\begin{tabular}{llcc}
%\toprule
%Dataset & Metric & AttnGAN & DM-GAN \\ \midrule
%CUB & FID$\downarrow$ & 23.98 & \textbf{16.09} \\
% & R-precision$\uparrow$ & 67.82$\pm$4.43 & \textbf{72.31$\pm$0.91}\\ \midrule
%COCO & FID$\downarrow$ & 35.49 & \textbf{32.64} \\
% & R-precision$\uparrow$ & 85.47$\pm$3.69 & \textbf{88.56$\pm$0.28}\\
%\bottomrule
%\end{tabular}
%\caption{Performance of FID and R-precision for AttnGAN \cite{Xu:2017wg} and our DM-GAN on the CUB and COCO datasets. The FID of AttnGAN is calculated from officially released weights. Lower is better for FID and higher is better for R-precision.}
%\label{table:FID}
%\end{table}

Overall, the results across multiple evaluation metrics on the CUHK-PEDES test dataset consistently indicate that our proposed MGD-GAN achieves better results against the other three methods, concerning both the visually authenticity and text-image consistency. 

Specifically, the Inception Score, Pose Score and Pose Variance of MGD-GAN and other methods are in Table \ref{table:eval_quality}. MGD-GAN achieves the best results across these three metrics. Compared with the state-of-art method on fine-grained text-to-image generation, the AttnGAN \cite{Xu:2018wgb}, MGD-GAN improved the Inception Score from 5.07 to 5.74, the Pose Score from 0.465 to 0.489 and the Pose Variance from 2.039 to 2.053. These results indicate that the proposed modules in MGD-GAN are promising directions for the fine-grained text-to-image generation.

As for the evaluation of whether the generated images are well-conditioned on the input text, the natural language generation evaluation results on the generated texts are listed in Table \ref{tab:eval_caption}. Our MGD-GAN achieves better results in terms of all the seven evaluation metrics. %This again verifies the effectiveness of exploring attention enhanced local-global discriminators.

\subsection{Qualitative evaluation}

Figure \ref{fig:compare} demonstrates the synthesized images produced by our MGD-GAN and three state-of-the-art models in the context of quality assessment. All samples are conditioned on text descriptions on CUHK-PEDES test set. Our MGD-GAN method produces pedestrians with a coherent structure and vivid details in most cases, comparing to the AttnGAN, the StackGAN and the Reed \etal. For convenience, we use $C_{ij}$ to refer the $i_{th}$ row and the $j_{th}$ column example.

\paragraph{Subjective analysis}

Due to the lack of attention mechanism designed for fine-grained image generation, StackGAN and Reed \etal  generate images with vague appearance ($C_{12}$, $C_{13}$) and inconsistent body structure ($C_{11}$, $C_{22}$). The AttnGAN method, which employs an attention mechanism through the generator, achieves better results. For example, the $C_{13}$ generated by AttnGAN looks much more realistic and has most details described by the text, including "white t-shirt" and "black pants". By comparing MGD-GAN with AttnGAN, we can see that MGD-GAN further improves many fine-grained details, such as the facial contour in $C_{21}$ and hairstyle in $C_{22}$, which depicts the femaleness. The result indicates the merit of leveraging fine-grained attention mechanism in both generator and discriminator collaboratively for generating semantically consistent images.

%---------------------------figure-------------------------

\begin{figure}[!t]
\centering
\includegraphics[width=\columnwidth]{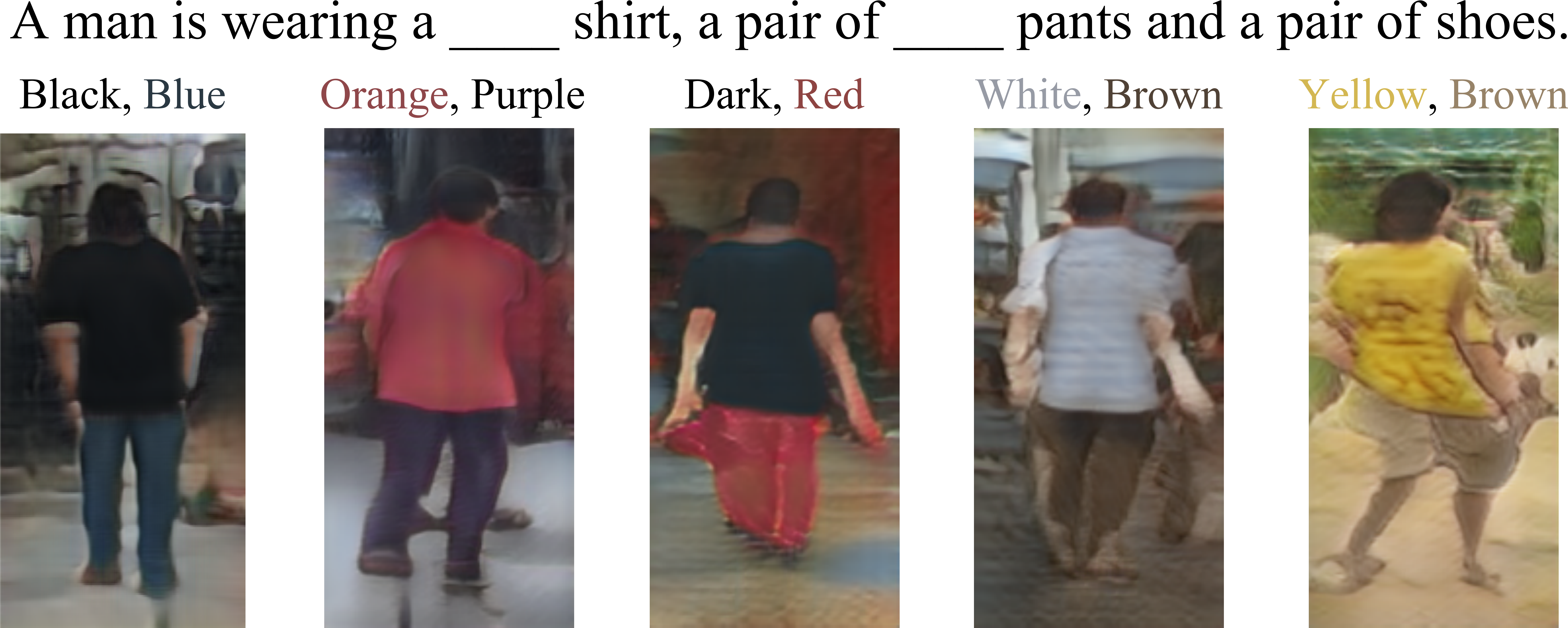}
\caption{
Controllable image generation by deliberately changing color attributes. %We additionally show that our model can synthesize visually authentic images with diversified appearances and human poses by deliberately changing the attribute.
}
\label{fig:colorChange}
\end{figure}

%------------------------------figure end-----------------------

\paragraph{Controllable image generation}

By altering the colors of wearings, Figure \ref{fig:colorChange} shows the controllable image generation results. Our model can generate pedestrians with fine-grained details authentic to the input text, and diversified poses and appearances. For example, there are pedestrians wearing long pants (the first columns) and short pants (the $5_{th}$ column), with hands in the pocket (the first column), hands putting in front (the second column) and hands stretching on both sides (the third column). %For example, there are pedestrians wearing long pants (the $2_{th}$ and the $6_{th}$ columns) and short pants (the first and the $10_{th}$ columns), with hands in the pocket (the $2_{th}$ column), hands putting in front (the $6_{th}$ column) and hands stretching on both sides (the $5_{th}$ column).

\paragraph{Multi-stage Refinement \& Attention visualization}

\begin{figure}[!t]
\centering
\includegraphics[width=\columnwidth]{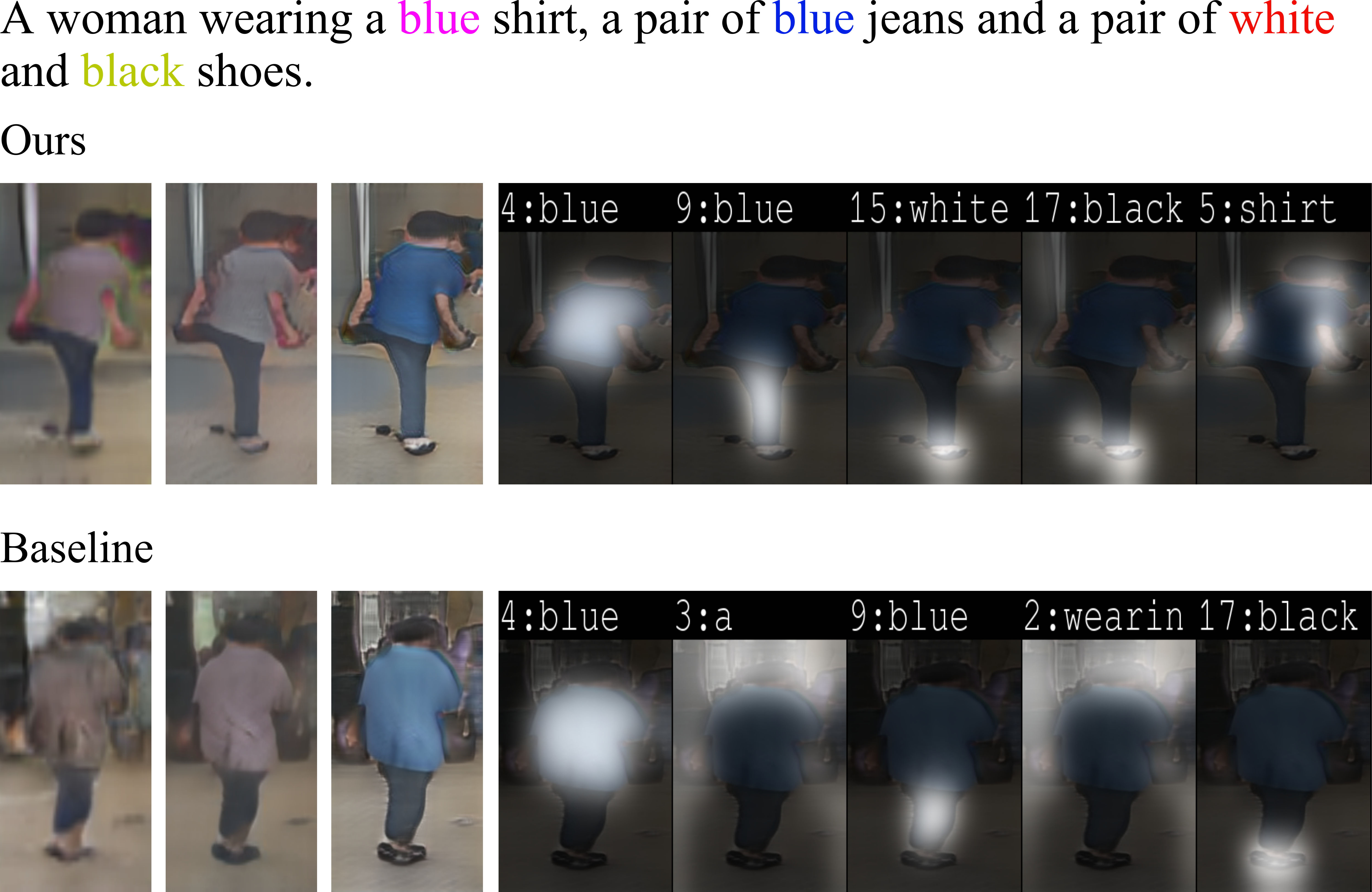}
\caption{
Comparison of the multi-stage generation results and the top 5 relevant words selected by attention module between the proposed method and the baseline method. %Our model can progressively refine the artifacts and missing attributes in images from the previous stages. The attended words of the generator in the MGD-GAN are more meaningful and more closely related to human appearance and structure.
}
\label{fig:attention}
\end{figure}

To better understand the effectiveness of our proposed modules, we visualize the multi-stage generation results and attention maps of the baseline method (without VISA-HPB and SCA-GD) and our method (Figure \ref{fig:attention}). It can be seen that 1) The MGD-GAN can generate more details than the baseline method, such as the hands and the "white and black shoes". The body structure is also better. 2) With a more balanced training of generator and discriminator, the attended words of the generator in our model are more meaningful and more closely related to human appearance and structure. 3) Our model can progressively refine the body structure, correct irregular artifacts, and remedy missing attributes in images from the previous stages.

%--------------------------------figure-------------------------

%----------------------------figure end-------------------------

%--------------------------------table-------------------------

\begin{table}[!t]
\centering
\setlength{\tabcolsep}{7.5pt}
\setlength\doublerulesep{0.5pt}
\begin{tabular}{l|cccc}

%\cline{1-9}
% <<<

\multicolumn{1}{c|}{Models} &    IS      &   PS    &    PV     \\
\hline \hline

%Inception Score            & 4.32 $\pm$ 0.194          & 4.79 $\pm$   0.184   & 5.07 $\pm$ 0.396    & 5.12 $\pm$   0.433        & 5.32 $\pm$ 0.357          & 5.59 $\pm$ 0.462                   & \textbf{5.74 $\pm$ 0.526}       \\
%\hline
%Pose Score  & 0.275   & 0.362    &  0.465   & 0.470   & 0.486   &\textbf{0.498}   & 0.489   \\
%\hline
%Pose Variance     &   1.935   & 1.945    & 2.039  & 1.979   & 1.981   & 2.018   & \textbf{2.053} 

Baseline            & 5.12 $\pm$   0.470          &    0.275        &    1.979                 \\
\hline
+HPD & 5.32 $\pm$ 0.357   & 0.486    &   1.981  \\
\hline
+HPD+VISA      &  5.59 $\pm$ 0.462   & \textbf{0.498}    & 2.018  
        \\
\hline
+HPD+VISA+SCA & \textbf{5.74 $\pm$ 0.526}    & 0.489    & \textbf{2.053} 
\\
\hline

\end{tabular}

\caption{
Ablation test of different modules in our MGD-GAN.
}
\label{table:ablation}
\end{table}

\subsection{Ablation study}

In order to verify the efficacy of different components and gain a better understanding of the network’s behavior, we conduct the ablation study by progressively removing control components. The control components are set up to HPD, VISA, and SCA, which stands for the human-part-based discriminator, the visual-semantic attention and the self-cross-attention respectively. The baseline method, denoted as BL for simplicity, is defined by removing the above modules from the MGD-GAN. The results are shown in Table \ref{table:ablation}.

Specifically, the BL achieves a slight performance boost over the AttnGAN by leveraging the VISA in generator instead of the original attention mechanism in AttnGAN. By equipping the human-part-based discriminator, the BL+PD model makes the relative improvement over the BL from 5.12 to 5.32 and 0.470 to 0.486 on Inception Score and Pose Score, respectively. It is worth noting that the VISA enhanced model (BL+PD+VISA) achieves the best performance on Pose Score metric, which verifies the effectiveness of VISA-HPD module in generating fine-grained details of different parts. Moreover, by additionally incorporating a self-cross-attended global discriminator, the MGD-GAN model (BL+PD+VISA+SCA) leads to the best performance on the Inception Score (12\% improvement over BL) and Pose Variance metrics.

%--------------------------------figure-------------------------
\begin{figure}[!t]
\centering
\includegraphics[width=\columnwidth]{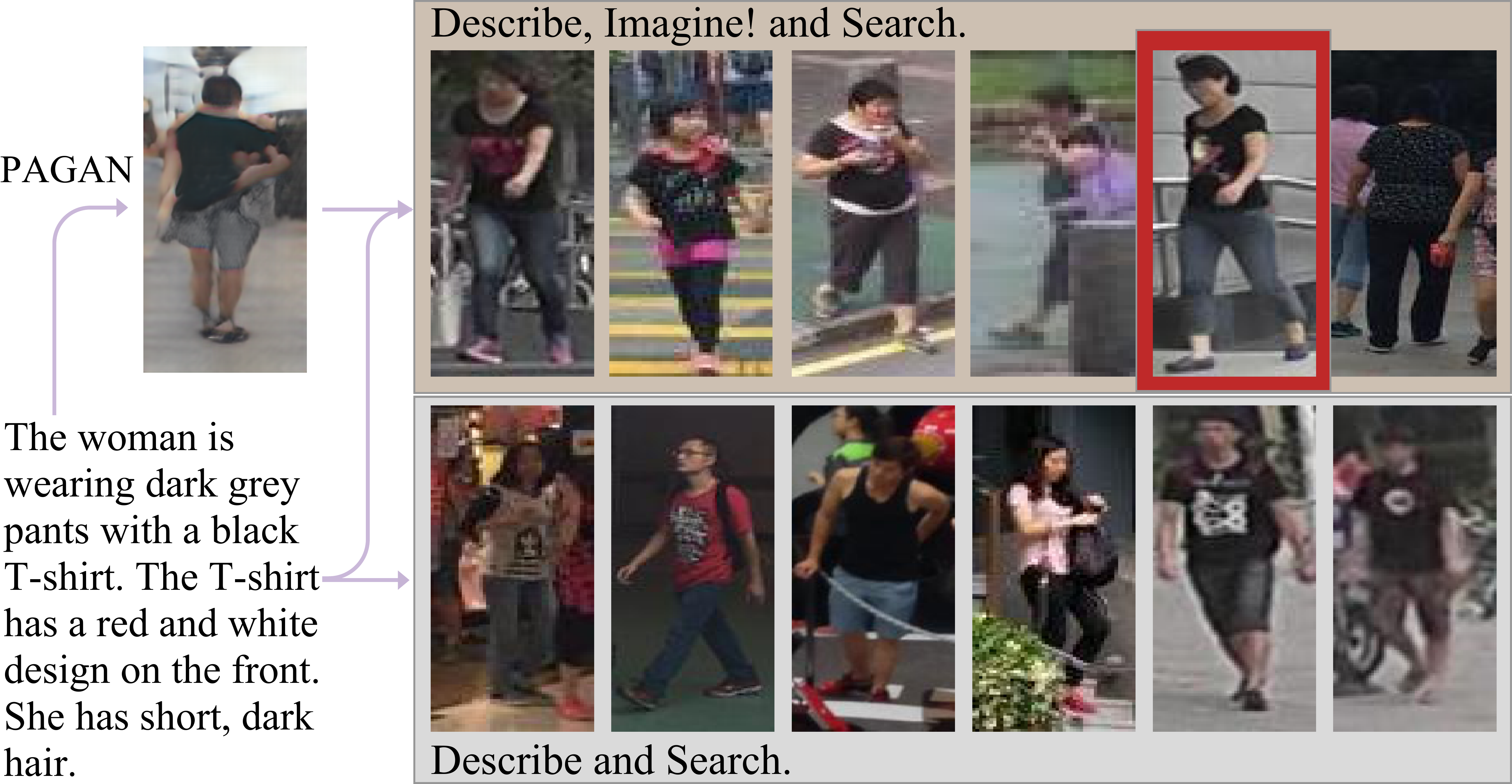}
\caption{
Results of MGD-GAN enhanced search and the baseline method. The image that is boxed of in red is the target image.
}
\label{fig:searchResults}
\end{figure}

%----------------------------figure end-------------------------

\subsection{Person Search Test}

%Text-to-pedestrian synthesis techniques have many real-world applications. For instance, in the automated surveillance scenario, when we only have the recorded words from witnesses and wishing to identify potential goals, text-to-pedestrian synthesis can help depict reasonable images based on vague descriptions. 
To verify that text-to-pedestrian can help bridge the gap between language and vision and thus promoting many real-world applications, we conduct a text-to-person search experiment the CUHK-PEDES dataset. 

%\begin{table}[!t]
%\centering
%\setlength{\tabcolsep}{6.5pt}
%\setlength\doublerulesep{0.5pt}
%\begin{tabular}{l|cc}
%%\toprule
%%Dataset & Metric & AttnGAN & DM-GAN \\ \midrule
%%CUB & FID$\downarrow$ & 23.98 & \textbf{16.09} \\
%% & R-precision$\uparrow$ & 67.82$\pm$4.43 & \textbf{72.31$\pm$0.91}\\ \midrule
%%COCO & FID$\downarrow$ & 35.49 & \textbf{32.64} \\
%% & R-precision$\uparrow$ & 85.47$\pm$3.69 & \textbf{88.56$\pm$0.28}\\
%
%Method  &   Baseline       & W. Fake Images \\ 
%\hline  \hline
%
%R@1 $\uparrow$  &  18.7   & \textbf{19.3} \\
%\hline
%R@5 $\uparrow$  &  46.1   & \textbf{47.4} \\
%\hline
%R@10 $\uparrow$  &  58.5   & \textbf{60.5} \\
%\hline
%Median $\downarrow$  &  7.0   & \textbf{6.0} \\
%\hline
%Average $\downarrow$  &  51.8   & \textbf{47.2} \\
%
%\hline
%\end{tabular}
%\caption{
%The results of generation enhanced person search. Higher is better for recall (R@1, R@5 and R@10) and lower is better for rank (median and average).
%}
%\label{table:search}
%
%\end{table}

\begin{table}[!t]
\centering
\setlength{\tabcolsep}{2.5pt}
\setlength\doublerulesep{0.5pt}
\begin{tabular}{l|ccccc}
%\toprule
%Dataset & Metric & AttnGAN & DM-GAN \\ \midrule
%CUB & FID$\downarrow$ & 23.98 & \textbf{16.09} \\
% & R-precision$\uparrow$ & 67.82$\pm$4.43 & \textbf{72.31$\pm$0.91}\\ \midrule
%COCO & FID$\downarrow$ & 35.49 & \textbf{32.64} \\
% & R-precision$\uparrow$ & 85.47$\pm$3.69 & \textbf{88.56$\pm$0.28}\\

Method  &  R@1 $\uparrow$   &  R@5 $\uparrow$ &  R@10 $\uparrow$ & Median $\downarrow$ &   Average $\downarrow$  \\ 
\hline  \hline

Baseline&  18.7   &  46.1 &  58.5 & 7.0 &   51.8 \\
\hline

W. Fake&  \textbf{19.3}   &  \textbf{47.4} &  \textbf{60.5} & \textbf{6.0} &   \textbf{47.2} \\
\hline

%  &    &  \\
%\hline
%  &    &  \\
%\hline
%  &    &  \\
%\hline
%  &    &  \\
%\hline
%  &    &  \\

\end{tabular}
\caption{
The results of generation enhanced person search. Higher is better for recall and lower is better for rank.
}
%(R@1, R@5 and R@10)
%(median and average)
\label{table:search}

\end{table}

We employ the off-the-shelf text-to-image search method named VSE++ \cite{faghri2018vse++} as the baseline method. We modify the baseline method by replacing the query from text only to text and an "imagination" picture. The "imagination" is a fake image generated by our MGD-GAN, which is pre-trained on the CUHK-PEDES training set. Original VSE++ contains a text encoder $Enc_{text}$ and a real image encoder $Enc_{real}$. We add a fake image encoder $Enc_{fake}$ which has the same structure as $Enc_{fake}$. $Enc_{real}$ and $Enc_{fake}$ are both pre-trained on ImageNet and fine-tuned during training but do not share weights. Instead of maximizing the similarity of the text representation and the real image representation, we replace the text representation with the joint representation of text and fake image. The joint representation is obtained using a fully connected layer.

The performance comparisons are summarized in Table \ref{table:search}. The consistent improvement over different metrics (Both recall and rank) indicate that our MGD-GAN can help bridge the gap across different modalities and thus enhancing text-based person search. Figure \ref{fig:searchResults} demonstrate a search example. Not only the MGD-GAN enhanced model found the target person image at top 5, but also the retrieved top six images are more relevant in details, such as the design on the front.

\section{Conclusion}

In this paper, we propose the MGD-GAN method for fine-grained text-to-pedestrian synthesis. We design a human-part-based discriminator equipped with the visual-semantic attention, which can be seen as a local discriminator functioning at the words-regions level, as well as a self-cross-attended global discriminator at the sentence-image level. We evaluate our model on the caption annotated person dataset CUHK-PEDES. The quantitative results across different metrics, including our proposed Pose Score \& Variance, and substantial qualitative results demonstrate the efficacy of our proposed MGD-GAN method.

\bibliography{refer}
\bibliographystyle{aaai}

\end{document}